\documentclass[11pt]{article}

\usepackage[margin=1in]{geometry}
\usepackage{microtype}

\usepackage{amsmath,amssymb,amsfonts}
\usepackage{algorithm}
\usepackage{algorithmic}

\usepackage{graphicx}
\usepackage{multirow}
\usepackage{booktabs}
\usepackage{float}

\usepackage[hidelinks]{hyperref}
\usepackage{xurl}
\usepackage{xcolor}
\usepackage{textcomp}
\usepackage{authblk}

\def\BibTeX{{\rm B\kern-.05em{\sc i\kern-.025em b}\kern-.08em
    T\kern-.1667em\lower.7ex\hbox{E}\kern-.125emX}}

\title{The Promise of Large Language Models in Digital Health:\\
Evidence from Sentiment Analysis in Online Health Communities%
\thanks{This work was supported in part by the National Institute for Health and Care Research Program Grant for Applied Research (NIHR PGfAR reference 202037, Effectiveness and cost-effectiveness of a digital social intervention for people with troublesome asthma promoted by primary care clinicians). The views expressed are those of the authors and not necessarily those of the National Health Service, National Institute for Health and Care Research, or the Department of Health and Social Care.}
}

\author[1]{Xiancheng Li\thanks{Corresponding author: \texttt{x.l.li@qmul.ac.uk}}}
\author[2]{Georgios D. Karampatakis}
\author[2]{Helen E. Wood}
\author[2]{Chris J. Griffiths}
\author[2]{Borislava Mihaylova}
\author[3]{Neil S. Coulson}
\author[4]{Alessio Pasinato}
\author[1]{Pietro Panzarasa}
\author[4]{Marco Viviani}
\author[2]{Anna De Simoni}

\affil[1]{School of Business and Management, Queen Mary University of London}
\affil[2]{Wolfson Institute of Population Health (WIPH), Queen Mary University of London}
\affil[3]{Department of Medicine, University of Nottingham}
\affil[4]{Department of Informatics, Systems, and Communication, University of Milano-Bicocca}

\date{} 

\begin{document}
\maketitle

\begin{abstract}
Digital health analytics face critical challenges nowadays. The sophisticated analysis of patient-generated health content, which contains complex emotional and medical contexts, requires scarce domain expertise, while traditional ML approaches are constrained by data shortage and privacy limitations in healthcare settings. Online Health Communities (OHCs) exemplify these challenges with mixed-sentiment posts, clinical terminology, and implicit emotional expressions that demand specialised knowledge for accurate Sentiment Analysis (SA). 

To address these challenges, this study explores how Large Language Models (LLMs) can integrate expert knowledge through in-context learning for SA, providing a scalable solution for sophisticated health data analysis. Specifically, we develop a structured codebook that systematically encodes expert interpretation guidelines, enabling LLMs to apply domain-specific knowledge through targeted prompting rather than extensive training.

Under this strategy, six GPT models validated alongside DeepSeek and LLaMA 3.1 are compared with pre-trained language models (BioBERT variants) and lexicon-based methods, using 400 expert-annotated posts from two health communities. LLMs achieve superior performance (81-89\% accuracy vs. 48-72\% for traditional methods) while demonstrating expert-level agreement (Fleiss' Kappa: 0.42-0.75). This high agreement, with no statistically significant difference (e.g., p$>$0.05) from inter-expert agreement levels, suggests knowledge integration beyond surface-level pattern recognition.

The consistent performance across diverse LLM models, supported by in-context learning, offers a promising solution for digital health analytics. Confidence calibration analysis shows that reasoning models provide reliable uncertainty estimates, enabling quality-controlled deployment in healthcare settings. This approach thus addresses the critical challenge of expert knowledge shortage in digital health research, enabling real-time, expert-quality analysis for patient monitoring, intervention assessment, and evidence-based health strategies. Our open-source implementation facilitates the immediate adoption of this methodology across diverse healthcare research contexts and tasks.

\end{abstract}


\section{Introduction}
\label{sec:introduction}

Over the past few years, \textit{Machine Learning} (ML) has demonstrated remarkable capabilities across various aspects of digital health, including diagnostics~\cite{awan2018machine}, personalised treatment~\cite{ahamed2018applying}, and patient monitoring~\cite{poalelungi2023advancing}. However, digital health analytics faces fundamental challenges that limit the widespread adoption of advanced computational approaches. Healthcare data is characterised by complex regulatory requirements, privacy constraints, and the need for domain-specific interpretation that goes beyond general-purpose analytics~\cite{seh2021machine}. These challenges are compounded by the scarcity of labelled training data due to privacy and governance restrictions, creating barriers to implementing sophisticated data analysis methods across digital health contexts~\cite{hossain2023natural}.

In the realm of digital health applications—from electronic health records to patient-generated content—text analysis presents unique interpretive challenges. Health-related text consistently involves specialised medical terminology, implicit clinical meanings, and context-dependent information that require domain expertise for accurate analysis~\cite{klug2024admission}. \textit{Online Health Communities} (OHCs) serve as a representative and particularly challenging use case, where patients sharing chronic condition experiences create posts containing complex medical information, personal narratives, and nuanced expressions that require sophisticated interpretation~\cite{carrillo2018feature}. The analytical challenges present in OHCs reflect broader difficulties in digital health text analysis, underscoring the need for innovative approaches to handle such complex content.

\textit{Sentiment Analysis (SA)} has been widely employed for quite some time to extract insights from health-related content, providing valuable opportunities to explore patient experiences, treatment outcomes, and the dynamics of OHCs as expressed through narrative text~\cite{denecke2023sentiment}. 
SA can reveal patterns related to patient engagement, treatment adherence, and emotional responses to interventions that would otherwise remain hidden in unstructured data~\cite{yang2016mining}. As illustrated in~\cite{carrillo2018feature,li2025understanding}, OHCs exemplify this potential, 
since SA can be employed to capture shifts in patient mood, treatment satisfaction, and peer support effectiveness over time, providing valuable indicators for healthcare quality improvement and patient monitoring. This is particularly relevant for chronic disease management, where patients often require sustained emotional support that clinicians may struggle to provide within limited consultation time, leading many to seek encouragement and understanding from peers who share similar experiences~\cite{chen2024you}. 
However, despite the recognised importance of SA in the above-mentioned scenario, existing approaches still face significant limitations. 
Most SA tools and models have been developed and validated on general social media data, with limited evaluation on the complex, domain-specific content found in healthcare contexts~\cite{pratiwi2024navigating,elbers2023sentiment}. Furthermore, traditional ML approaches require extensive labelled datasets for effective performance, yet such data remains scarce in healthcare due to privacy constraints and the high cost of expert annotation~\cite{alanazi2022using,peiffer2020machine}.

\textit{Large Language Models} (LLMs) such as GPT, LLaMA, and DeepSeek, represent today a transformative opportunity to address both data scarcity and domain expertise challenges simultaneously in modern digital health analytics. Through extensive pre-training on diverse text corpora, including medical literature, LLMs possess inherent domain knowledge that can be activated and directed through structured \textit{prompting} approaches~\cite{chang2024survey}. Crucially, LLMs demonstrate the ability to integrate expert-derived guidelines and perform sophisticated analysis tasks without requiring extensive labelled training data through in-context learning, offering a pathway to scale expert knowledge in healthcare contexts where such expertise is traditionally scarce and expensive~\cite{zhong2023can,amin2023wide}. This capability represents a fundamental shift from traditional data-intensive approaches to knowledge-guided methodologies, potentially democratising access to sophisticated health data interpretation across diverse digital health applications.

In light of the above, leveraging LLMs for the task of SA appears to be a highly promising direction for healthcare analytics. While this potential is significant, recent studies have primarily demonstrated their capabilities in general-domain settings~\cite{zhong2023can,amin2023wide}, and there remains a notable gap in understanding how these models perform when applied to health-related content, where specialised terminology, contextual subtlety, and emotional complexity demand more nuanced interpretation. Most critically, the fundamental question of whether LLMs can effectively integrate domain-specific expert knowledge to achieve sophisticated analysis without extensive training data requirements has not been systematically evaluated in healthcare settings, where such expertise is traditionally scarce and expensive to scale~\cite{sushil2024comparative}. In addition, for practical healthcare deployment, sophisticated analysis capabilities alone are insufficient—healthcare applications require a reliable assessment of when automated predictions can be trusted and when expert review is necessary. A natural approach to address this challenge is to leverage LLMs' capacity for self-assessment, where models evaluate their own prediction confidence~\cite{gligoric2024can}. While LLMs can indeed be prompted to provide confidence scores alongside their predictions, the meaningfulness and reliability of these uncertainty estimates in healthcare contexts remain largely unexplored~\cite{zhou2024relying}. 

A further deployment concern is confidence calibration—whether a model’s predicted confidence aligns with its actual accuracy. Traditional performance metrics (e.g., accuracy, precision, recall) do not capture prediction reliability, yet in many contexts, understanding when a model is uncertain can be as important as the prediction itself. Therefore, systematic assessment of calibration is essential to determine whether LLMs can provide reliable uncertainty quantification. Such capability would enable quality-controlled deployment strategies in which high-confidence predictions are processed automatically while uncertain cases receive expert oversight, optimising the balance between automation efficiency and quality assurance in healthcare applications.

To tackle the above-mentioned issues, this study specifically aims to evaluate the potential of LLMs to integrate expert knowledge for SA in digital health contexts. To demonstrate the effectiveness of expert knowledge integration, we develop a structured codebook that encodes domain-specific interpretation guidelines, enabling LLMs to apply expert-derived rules for consistent sentiment classification, carefully taking into account the confidence calibration analysis. We focus primarily on GPT models due to their widespread accessibility and well-documented performance~\cite{rathje2024gpt,gilardi2023chatgpt}, whilst also evaluating other state-of-the-art open-source LLMs including DeepSeek and LLaMA, to establish the generalisability of our findings across different model architectures. The inclusion of open-source models is particularly important for healthcare applications, as they enable local deployment that can better ensure data security and privacy compliance when working with confidential medical data. By comparing LLM performance against traditional ML and lexicon-based approaches, we demonstrate how expert knowledge integration can achieve sophisticated analysis without extensive training data requirements, establishing a new paradigm for scalable, expert-quality analysis in digital health applications. To facilitate broader adoption, we provide a complete open-source implementation that enables immediate application of our solution across diverse healthcare research contexts and expert annotation tasks in healthcare beyond SA.

\subsection{Related Work}

The application of advanced \textit{Natural Language Processing (NLP)} and, more in general, \textit{Artificial Intelligence (AI)} techniques, particularly LLMs, in digital health studies has gained increasing attention in recent years. Within NLP, SA is particularly relevant for digital health, as it supports the systematic understanding of patient experiences and emotional states from unstructured text.

Traditional approaches to SA span three families:
(i) lexicon-based methods~\cite{taboada2011lexicon};
(ii) classical ML classifiers~\cite{pang2002thumbs}; and
(iii) transformer-based architectures (e.g., BERT)
~\cite{devlin2019bert}.

In OHCs, these approaches have been used to analyse symptom-focused discussions~\cite{xiang2023study}, to monitor emotional trajectories in patient interactions~\cite{li2025understanding}, and to detect mental health concerns or at-risk users in forum settings~\cite{kim2023understanding}.
However, health-related text introduces additional challenges, including domain shift between general and medical language, frequent abbreviations and jargon, implicit or mixed sentiment within single messages, and the scarcity of high-quality annotated datasets ~\cite{villanueva2025sentiment}. These factors increase the risk of domain shift when models are applied outside their original training distribution and can degrade performance in OHCs compared with general social media.

Recent advances in large language models (LLMs) have introduced new possibilities for SA through zero- and few-shot in-context learning, reducing reliance on large task-specific training datasets while improving adaptability across domains. This growing body of work motivates further examination of how LLMs perform in both general-domain and health-specific SA tasks. Analyses have shown that LLMs achieve similar or even better performance compared to other approaches when applied to SA tasks using data from Reddit, Twitter (now X), and Amazon Review~\cite{zhong2023can,amin2023wide,belal2023leveraging}. These general-domain tasks often focus on product reviews or social commentary, where sentiment tends to be explicit and polarity is easier to detect. SA has also been applied in health-related contexts. For example, Pratiwi et al.~\cite{pratiwi2024navigating} used a dictionary-based approach to analyse asthma-related discussions on Twitter, while Elbers et al.~\cite{elbers2023sentiment} applied sentiment scoring to clinical notes from lung cancer patients to examine emotional trends around diagnosis and treatment. However, such applications typically focus on method demonstration rather than systematic performance evaluation, and often fail to address the inherent complexity of health-related text that requires domain-specific interpretation.

Recent comparative studies have begun to address these limitations by systematically evaluating different model architectures on health-specific datasets. Elmitwalli and Mehegan~\cite{elmitwalli2024sentiment} conducted a comprehensive comparison of lexicon-based, deep learning, and pre-trained language models using both general-purpose datasets (IMDB, Sentiment140) and domain-specific medical content (COP9-related tweets), demonstrating that pre-trained models consistently outperformed traditional approaches. Their findings showed GPT-3 achieving superior performance on the medical datasets, suggesting strong zero-shot capabilities in health-specific contexts. However, Zhang et al.~\cite{zhang2020sentiment} reported contrasting results when evaluating fine-tuned GPT and BERT models on HPV vaccine-related Twitter content, with BERT outperforming GPT in their specific task setting. These mixed findings highlight a critical challenge in health-related SA: model performance appears highly dependent on dataset characteristics, task formulation, and fine-tuning strategies, making it difficult to establish generalisable conclusions about optimal approaches. Furthermore, these studies remain constrained to short-form Twitter content with explicit medical terminology, leaving significant gaps in understanding how these models perform on more complex, narrative-rich health communications.

In contrast, OHCs represent a fundamentally different environment for SA, where users engage in extended discussions about chronic conditions, treatment experiences, and peer support. Unlike the brief, topic-specific posts typical of Twitter, OHC participants—predominantly patients, caregivers, or family members—create detailed personal narratives that often exhibit shifting emotional tones within single messages. These posts may simultaneously express gratitude for community support, anxiety about upcoming treatments, and frustration with healthcare systems, creating complex sentiment patterns that require sophisticated interpretation. The personal, experiential nature of OHC content, combined with the emotional complexity inherent in chronic disease management, presents unique challenges for automated SA that extend well beyond the capabilities of approaches designed for general social media platforms.

\section{Methods}
\label{sec:methods}

This section describes the methodology adopted to perform SA on OHC data. We first present the datasets used in the study and the human annotation process. Next, we introduce the SA models considered for comparative evaluation. Finally, we describe the performance assessment framework and the metrics used to compare the models.

\subsection{OHC Datasets and Human Annotation}
\label{subsec: annotation}

As described in our previous work~\cite{joglekar2018online}, data were collected by \textit{HealthUnlocked},\footnote{\url{https://healthunlocked.com/}} the platform provider of the \textit{Asthma} + \textit{Lung} UK (AUK) and \textit{British Lung Foundation} (BLF) OHCs. Only posts that were shared publicly were collected and analysed. Our datasets were stored and analysed on a secure server held by Queen Mary University of London. Anonymised user identifiers (IDs) were provided by HealthUnlocked, and no demographic information was available. Our data contains three types of posts, denoted as: $(i)$ $level$-0 $posts$ (i.e., posts starting new threads), $(ii)$ $level$-1 $replies$ (i.e., replies to the level-0 posts), and $(iii)$ $level$-2+ $replies$ (i.e., replies to level-1 replies and above). The data sets consisted of 12,453 posts since 2006 for AUK, and of 367,787 posts since 2012 for BLF. We used stratified random sampling without replacement to select 200 posts from each dataset, maintaining the original distribution of post types.

Five expert \textit{human annotators} independently added \textit{sentiment labels} (i.e., positive, neutral, or negative) to the two sets of 200 random posts from each OHC. All annotators hold PhD degrees: Annotators 2 and 5 are native English speakers, Annotators 2, 4 and 5 specialise in qualitative analysis in healthcare research, Annotator 4 is a General Practitioner with clinical experience, and Annotator 1 is a data scientist with extensive experience in OHC data analysis. 

\subsubsection{Structured Codebook}
\label{sec:codebook}
We developed a structured codebook to operationalise the sentiment schema used in this study. It consolidates (i) the label set with definitions and decision rules, (ii) inclusion/exclusion criteria and guidance for borderline or implicit cases, and (iii) prototypical examples with brief rationales. The codebook was created through a rigorous consensus process: annotators first independently labelled a 400-post pilot set (distinct from the final dataset), then disagreements were reviewed in multiple meetings to harmonise interpretations and formalise rules. The final version includes comprehensive sentiment definitions, instructions for handling implicit sentiment, examples with justifications, and specific rules for ambiguous or mixed-sentiment cases. We used the codebook both to train annotators and standardise labelling—supporting inter-annotator agreement and quality control—and to construct LLM prompts: zero-shot prompts comprised task instructions plus the codebook’s label definitions and decision rules (no examples), while few-shot prompts appended a small set of representative examples with gold labels. This alignment ensures that humans and models follow the same expert-derived labelling logic and facilitates the integration of domain-specific knowledge into automated analysis.

\subsection{Sentiment Analysis Models}

Traditional SA approaches have primarily relied on two classes of methods: \textit{lexicon-based} techniques that use predefined word lists categorised by sentiment polarity, and \textit{semantic-based} techniques based on supervised ML where classifiers are trained on labelled datasets~\cite{turney2002thumbs}. Whilst lexicon-based methods offer simplicity and interpretability, they often lack contextual understanding, and supervised approaches require extensive annotated training data that may not be available in specialised domains. To evaluate the potential of LLMs for expert knowledge integration in health-specific SA, we compared their performance against these traditional approaches. We employed the following three categories of models to generate sentiment labels.

\subsubsection{Lexicon-based models}
 
We used three widely-adopted lexicon-based models: \textit{Valence Aware Dictionary for Sentiment Reasoning} (VADER)~\cite{hutto2014vader}, \textit{TextBlob}~\cite{loria2018textblob}, and \textit{SentiWordNet}~\cite{ baccianella2010sentiwordnet}. These models were implemented using standard Python libraries and serve as established baseline approaches for SA tasks\footnote{%
\begin{tabular}{@{}p{2.3cm}p{\dimexpr\linewidth-2.3cm\relax}@{}}
VADER:        & \url{https://github.com/cjhutto/vaderSentiment} \\
TextBlob:     & \url{https://textblob.readthedocs.io/en/dev/} \\
SentiWordNet: & \url{https://www.nltk.org/howto/sentiwordnet.html}
\end{tabular}}. Lexicon-based models generate a compound score indicating the sentiment polarity of each post based on predefined sentiment lexicons. The compound score ranges from $-1$ to $1$, where negative and positive values correspond to negative and positive sentiment, respectively. We classified posts with compound scores between $-0.1$ and $0.1$ as neutral sentiment.

\subsubsection{Fine-tuned BioBERT models}

BioBERT is a domain-specific language representation model pre-trained on large-scale biomedical data~\cite{lee2020biobert}. We selected it as a representative example of pre-trained language models requiring supervised fine-tuning, as such models have demonstrated promising performance in prior SA tasks involving health-related texts, making BioBERT a suitable baseline for comparison with LLMs and lexicon-based approaches. We fine-tuned BioBERT with four datasets:
\begin{enumerate}
    \item 143,903 Covid-19 tweets~\cite{chakraborty2021sentiment}, where sentiment labels were produced by NLTK, a Python library for NLP;
    \item 50,333 tweets from SemEval-2017~\cite{rosenthal2019semeval}, where sentiment labels were produced by annotators recruited through the CrowdFlower platform;
    \item A combination of Covid-19 and SemEval-2017 tweets;
    \item A balanced sub-sample of Covid-19 and SemEval-2017 tweets containing 48k tweets where a random sample of 16k tweets was selected for each sentiment label. 
\end{enumerate}

In the remainder of the paper, the fine-tuned BioBERT models using the four datasets are referred to as BioBERT-Covid, BioBERT-SemEval, BioBERT-Combine, and BioBERT-Balance, respectively.

\subsubsection{Large Language Models}

LLMs represent a paradigm shift in NLP, offering the ability to perform complex text analysis tasks through in-context learning via zero-shot and few-shot learning without requiring extensive task-specific training data. Our method implements expert knowledge integration by embedding domain-specific interpretation guidelines and labelling rules directly into structured prompts, derived from our expert-developed codebook. We evaluated multiple LLM architectures to demonstrate the generalisability of this expert knowledge integration approach across different model families.

\textbf{GPT Models:} We selected GPT models as our primary LLM evaluation target due to their widespread accessibility, well-documented performance, and user-friendly API interface that enables broader adoption in healthcare research settings. In this study, we used OpenAI's API to query the latest models GPT-4.1, GPT-o3, and their corresponding mini variants (i.e., GPT-4.1-mini, GPT-o3-mini), which offer different computational efficiency trade-offs whilst maintaining strong performance capabilities.

We evaluated both zero-shot and few-shot learning approaches to assess how effectively this structured prompting methodology could transfer expert knowledge without requiring extensive training data. For zero-shot learning, we used prompts that incorporated the expert-derived codebook guidelines and labelling rules without providing specific examples. For few-shot learning, we enhanced the structured prompts with a few carefully selected examples, referred to as GPT-4.1-fs, GPT-o3-fs, GPT-4.1-mini-fs, and GPT-o3-mini-fs, respectively. The few-shot prompts included the same codebook guidelines plus specific examples to demonstrate the application of labelling rules. Both approaches demonstrate how domain-specific knowledge can be systematically transferred to LLMs through structured prompting approaches.

\textbf{Other LLM Models:} To establish the generalisability of our findings across different LLM architectures, we evaluated recently released models including DeepSeek (versions R1 and V3) and LLaMA 3.1 (70B and 405B). These open-source alternatives are particularly important for healthcare applications, as they enable local deployment that can better ensure data security and privacy compliance when working with confidential medical data. The performance of these models is summarised in the text to complement our main findings and demonstrate the robustness of the LLM approach across different architectures, while detailed graphical analysis focuses on the GPT model family.

To ensure reproducibility and facilitate broader adoption, we provide open-source code implementing our structured codebook approach on GitHub\footnote{\url{https://github.com/XianchengLI/sentiment-analysis-llm-health}}. The repository includes the complete implementation code, evaluation scripts, the codebook, zero-shot and few-shot prompts with embedded codebook rules, and example usage demonstrating the methodology.
    
\subsection{Performance Assessment}
\label{subsec: perf_assess}
Our comparative evaluation covers five aspects: label distribution analysis, inter-annotator agreement among experts, agreement of LLMs with respect to individual annotators, detailed performance against the majority label, and confidence calibration.

\begin{enumerate}

\item \textit{Label distribution analysis.} For each dataset, we computed the marginal frequency of Positive, Negative and Neutral labels produced by human annotators and models through visualisation. This analysis is descriptive and aims to characterise broad tendencies in assigned labels across human annotators and models.
\item \textit{Inter-annotator agreement.} Among the five expert annotators who independently provided their sentiment labels for the random samples---as previously introduced in Section \ref{subsec: annotation}---was assessed using \textit{Fleiss' kappa} coefficient (denoted as $\kappa$), which reduces to \textit{Cohen's kappa} coefficient for pairwise comparisons~\cite{fleiss1971measuring}, to evaluate the extent to which the five annotators agreed with each other. Statistical significance of kappa coefficients was evaluated using asymptotic $z$-tests against the null hypothesis of $\kappa = 0$, with significance level set at $\alpha = 0.05$. Following established practices~\cite{rathje2024gpt,carrillo2018feature}, we assigned a new label to each post based on the majority consensus (referred to hereafter as the \textit{majority label});
\item \textit{LLM–expert agreement.}, To test whether LLMs achieve agreement levels comparable to human experts, we compared the distribution of LLM–human pairwise $\kappa$ values to the distribution of human–human pairwise $\kappa$ values using the Mann–Whitney~U test~\cite{nachar2008mann} (non-parametric, no normality assumption; $\alpha=0.05$). Human–human agreement comprised all 10 annotator pairs ($n=10$), and LLM–human agreement comprised all model–annotator pairs across models ($n=40$). A non-significant result ($p>0.05$) indicates that LLM–human agreement shows no statistically significant difference from expert-level agreement.
\item \textit{Performance against the majority label.} To evaluate the \textit{effectiveness of expert knowledge integration} across the three categories of models, we employed a multi-faceted assessment approach. First, we used agreement metrics to report the level of agreement between different models and the expert annotators, with high agreement indicating successful knowledge transfer. We then considered the majority label as the benchmark and computed the \textit{accuracy} of each model (i.e., the percentage of correct labels). Additionally, we calculated \textit{precision}, \textit{recall}, and \textit{F1-score} for each sentiment category to provide a comprehensive performance evaluation~\cite{powers2020evaluation};
\item \textit{Confidence estimation and calibration.} To evaluate \textit{prediction reliability} for practical healthcare deployment, we implemented confidence estimation for selected GPT models capable of providing uncertainty quantification. Confidence scores were obtained by requesting models to report their certainty level ($[0-1]$ scale) alongside predictions, where $0$ indicates complete uncertainty and $1$ indicates complete certainty. Detailed confidence-enabled prompts and rules are available in our open-source repository. Confidence calibration was assessed using reliability diagrams plotting predicted confidence against actual accuracy across confidence bins. We also analysed confidence score distributions to evaluate whether models produce informative estimates across the full range or concentrate in narrow bands that limit utility for quality assessment. This analysis assesses model suitability for applications requiring uncertainty quantification and prediction reliability evaluation.
\end{enumerate}

\section{Results}
\label{sec:results}

This section is dedicated to presenting and discussing the results of the comparative evaluation with respect to the five key points outlined previously in Section \ref{subsec: perf_assess}.

\begin{figure}[!t]
\centerline{\includegraphics[width=\columnwidth]{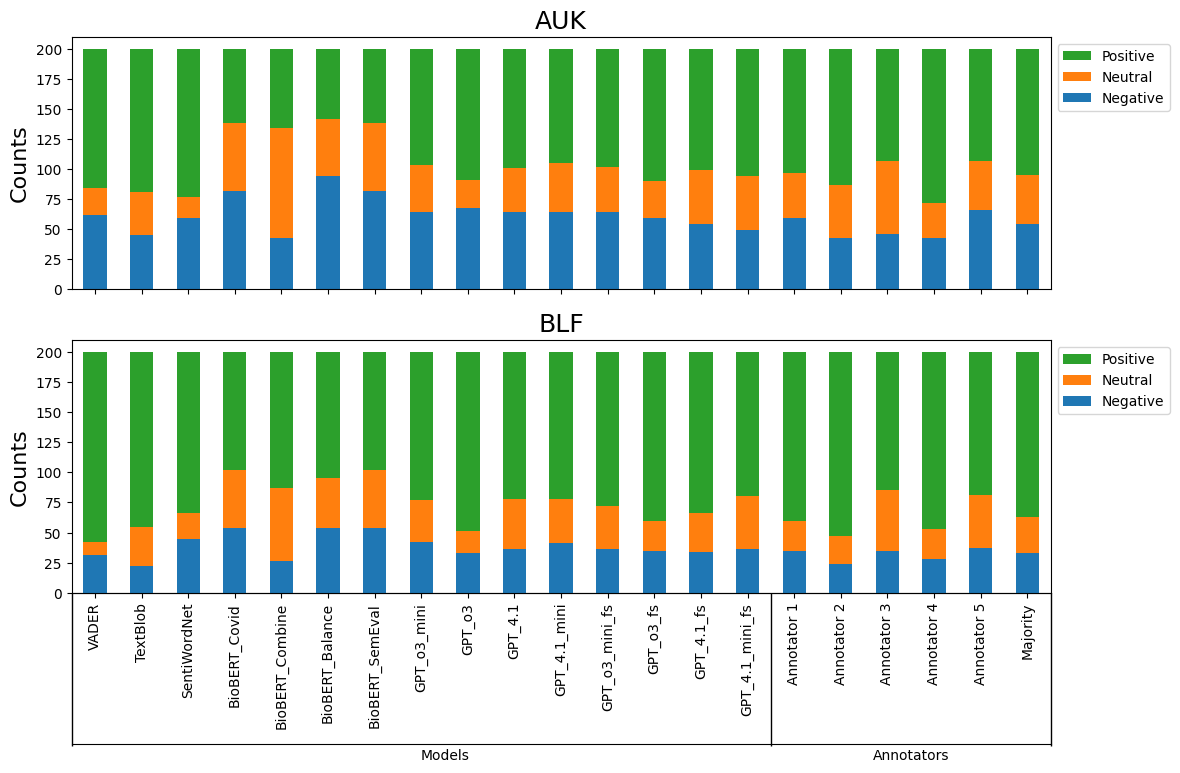}}
\caption{Distribution of labels given by each of the annotators and models to selected posts from the AUK OHC (upper panel) and the BLF OHC (lower panel).}
\label{fig:distribution}
\end{figure}

Concerning the first key point, Figure~\ref{fig:distribution} shows the distribution of sentiment labels produced by our selected models and expert annotators. Across both datasets, positive labels were more frequently assigned than negative or neutral labels by both human experts and most models. DeepSeek and LLaMA demonstrated similar distributional patterns, with positive sentiment being the most commonly assigned label across both datasets.

\begin{figure}[H]
\centerline{\includegraphics[width=0.85\columnwidth]{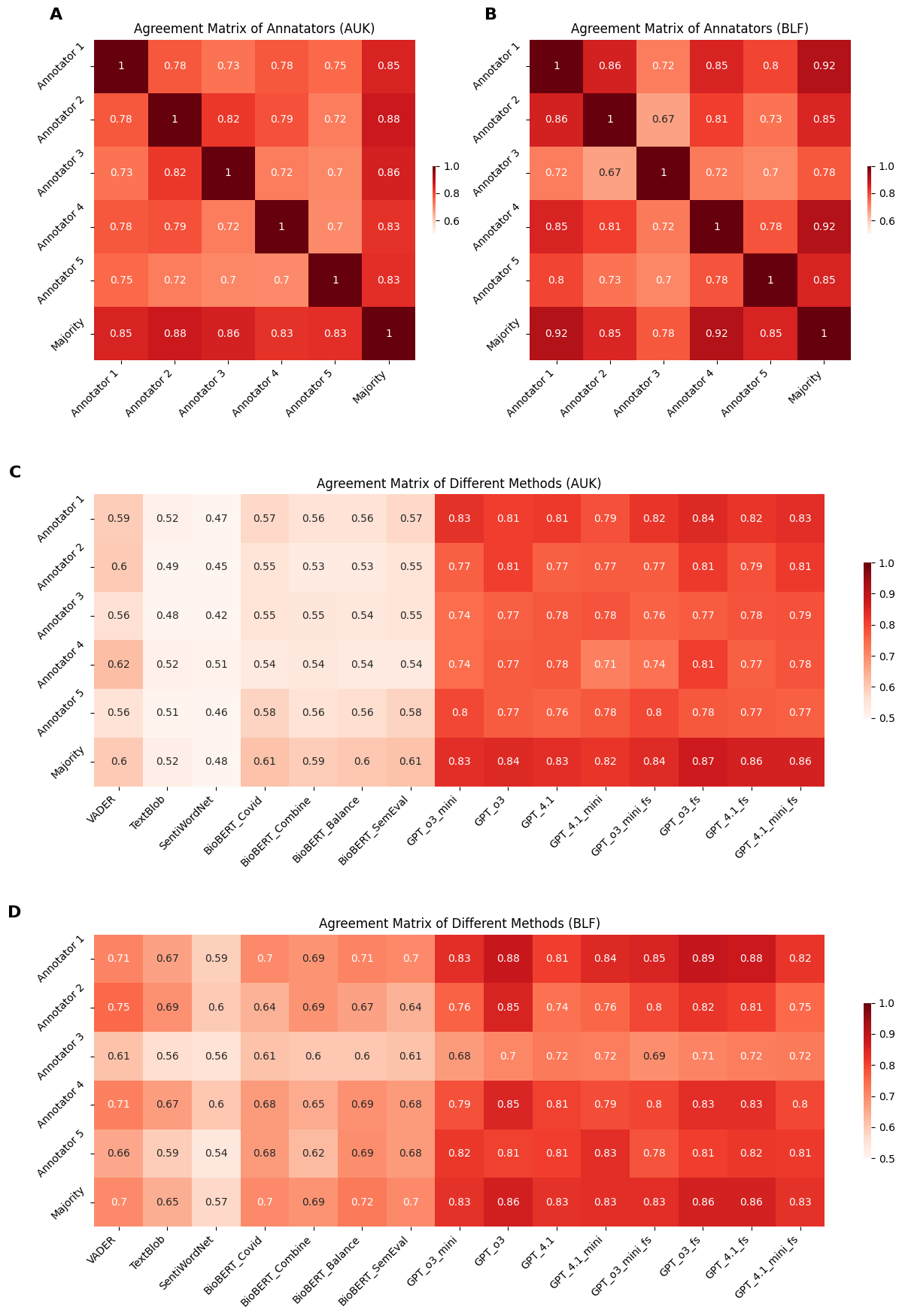}}
\caption{Agreement Matrix of annotators and all models. Agreements are measured by the ratio of posts with the same labels assigned by two annotators or models. Panels A and B show the agreement among annotators. Panels C and D show the agreement between annotators and models.}
\label{fig:agreement}
\end{figure}

Concerning the second key point, Figures~\ref{fig:agreement}A and~\ref{fig:agreement}B present the agreement matrix among expert annotators. Majority consensus labels could be determined for all posts in both datasets, reflecting the effectiveness of our expert annotation process. For the AUK sample, inter-annotator agreement ($\kappa$) ranged from 0.49 to 0.71 with an overall $\kappa$ of 0.59, indicating moderate agreement among experts. The highest agreement was between Annotators 1 and 2 with 82\% agreement ($\kappa = 0.71$). Individual annotator agreement with the majority consensus ranged from 0.70 to 0.79. For the BLF sample, inter-annotator $\kappa$ values ranged from 0.33 to 0.67 with an overall $\kappa$ of 0.52, also indicating moderate agreement. The highest agreement was between Annotators 1 and 2 with 86\% agreement ($\kappa = 0.67$). Agreement with majority consensus ranged from 0.58 to 0.82. All $\kappa$ values were statistically significant ($p$ $<$ 0.001), confirming genuine agreement beyond chance levels.

Concerning the third key point, Figure~\ref{fig:agreement}C and~\ref{fig:agreement}D present the agreement of the selected models with individual annotators. Although the agreement varies among different annotators, LLM models have consistently higher agreement with annotators than the other models. The agreement between GPT models and annotators ranged from 68\% to 89\% with $\kappa$ ranging from 0.42 to 0.75, which aligns with the degree of agreement among annotators themselves. Mann-Whitney U test revealed no significant difference between LLM-human agreement scores and human-human agreement scores ($p$ = 0.68 in AUK and $p$ = 0.10 in BLF), confirming that LLMs achieved genuine expert-level agreement rather than systematic bias.
This level of agreement indicates successful integration of expert knowledge, as the models achieved human-level consistency in sentiment classification without requiring extensive training data. DeepSeek and LLaMA demonstrated similar performance, with agreement ranging from 69\% to 92\% with $\kappa$ ranging from 0.45 to 0.82, confirming the robustness of expert knowledge integration across different LLM architectures.

We then selected one model at a time as an additional annotator and calculated the overall $\kappa$ among all annotators including the selected model (Table~\ref{tab:overall_kappa}). When adding labels from any LLM, the overall $\kappa$ increased from baseline values (0.586 for AUK, 0.524 for BLF). In contrast, the overall $\kappa$ decreased when adding labels from traditional approaches (lexicon-based and BioBERT models). Similar patterns were observed with DeepSeek and LLaMA models, which also increased overall agreement when included as additional annotators. These patterns suggest that LLM-generated labels showed no statistically significant difference from expert annotations in agreement trends.

\begin{table}[H]
\caption{Overall $\kappa$ calculated after adding labels from the selected model to those provided by the human annotators. Overall $\kappa$ values higher than the annotators only value are in bold.}
\label{tab:overall_kappa}
\centering
\begin{tabular}{|l|c|c|}
\hline
Source of labels added & Overall $\kappa$ (AUK) & Overall $\kappa$ (BLF) \\
\hline
Annotators only (Baseline) & 0.586 & 0.524 \\
\hline
VADER & 0.409 & 0.449 \\
TextBlob & 0.445 & 0.427 \\
SentiWordNet & 0.419 & 0.398 \\
BioBERT-Covid & 0.498 & 0.485 \\
BioBERT-SemEval & 0.498 & 0.485 \\
BioBERT-Combine & 0.490 & 0.465 \\
BioBERT-Balance & 0.491 & 0.487 \\
\hline
GPT-o3 & \textbf{0.604} & \textbf{0.549} \\
GPT-o3-mini & \textbf{0.604} & \textbf{0.540} \\
GPT-4.1 & \textbf{0.604} & \textbf{0.542} \\
GPT-4.1-mini & \textbf{0.598} & \textbf{0.550} \\
GPT-o3-mini-fs & \textbf{0.604} & \textbf{0.542} \\
GPT-o3-fs & \textbf{0.614} & \textbf{0.552} \\
GPT-4.1-fs & \textbf{0.608} & \textbf{0.556} \\
GPT-4.1-mini-fs & \textbf{0.612} & \textbf{0.546} \\
\hline
\end{tabular}
\end{table}

\begin{figure}[!t]
\centerline{\includegraphics[width=\columnwidth]{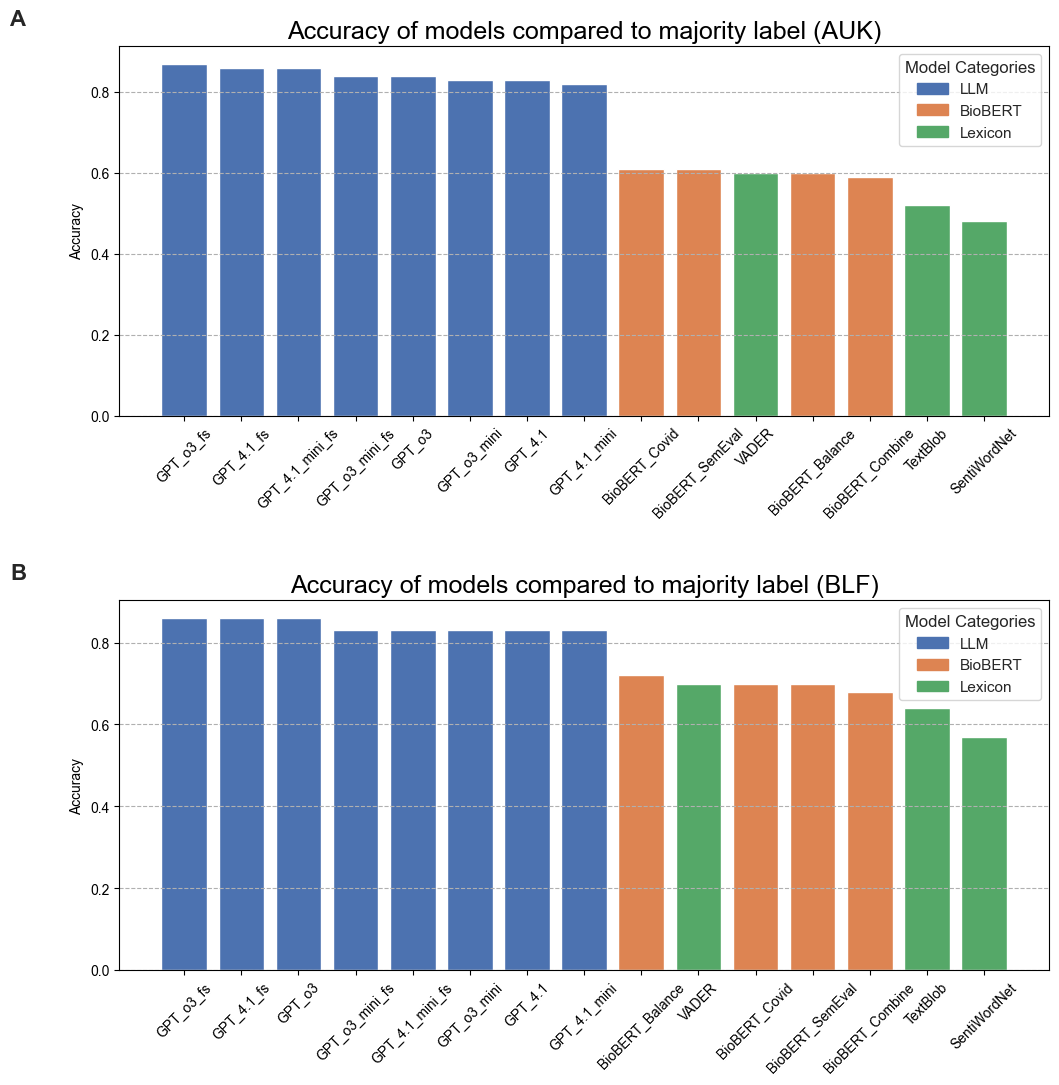}}
\caption{Accuracy of models. Accuracy is measured by the ratio of posts with same labels assigned by the model and majority label.}
\label{fig:accuracy}
\end{figure}

Concerning the fourth key point, Figure~\ref{fig:accuracy}A presents the accuracy of different models compared to the majority labels based on the AUK sample. LLMs demonstrated superior performance, with GPT-o3-fs achieving the highest accuracy (87\%), whilst other GPT variants achieved accuracy between 82\% and 86\%. Traditional approaches performed considerably lower, with BioBERT models (48\%-61\%) slightly outperforming lexicon-based models, among which VADER achieved the best accuracy. For the BLF sample (Figure~\ref{fig:accuracy}B), similar patterns emerged with GPT-o3-fs achieving 86\% accuracy and other GPT variants ranging from 83\% to 86\%. Traditional approaches again performed lower (57\%-72\%). Across both datasets, LLMs consistently achieved the highest accuracy, with GPT-o3-fs being the most accurate overall. 

Analysis of the accuracy results also reveals differences between GPT model generations and zero-shot versus few-shot approaches. As expected, full-size models (GPT-4.1 and GPT-o3) performed better than mini variants (GPT-4.1-mini and GPT-o3-mini). Few-shot learning demonstrated the potential for performance improvement, with GPT-4.1 showing particular sensitivity to few-shot examples, increasing accuracy from 83\% to 86\% on both datasets. The inclusion of expert-derived examples in few-shot prompts resulted in accuracy improvements of up to 3\%. Additional LLMs achieved comparable performance, with accuracy ranging from 81\% to 88\% on AUK and 82\% to 89\% on BLF samples. Notably, LLaMA 3.1 405B consistently achieved the highest accuracy across both datasets with strong zero-shot capabilities, whilst other LLMs demonstrated performance comparable to GPT models.

We then measured the performance using precision, recall, and F1-score across different sentiment labels (negative, neutral, positive). Table~\ref{tab:performance} presents the results on both datasets. Whilst the model with the highest F1-score varied across different sentiment labels, the best performers consistently belonged to the LLM category. As F1-score balances precision and recall, our results demonstrate that LLMs consistently outperform traditional approaches across all three sentiment categories. 

Table~\ref{tab:best_performance} presents a comprehensive comparison of the best-performing models of each category within each dataset. The results clearly demonstrate the superior performance of LLMs compared to traditional approaches. LLaMA 3.1-405B-fs achieved the highest performance across the majority of metrics, securing the best F1-scores for five out of six sentiment categories across both datasets. DeepSeek V3-fs also demonstrated strong performance, particularly excelling in negative sentiment recall. In contrast, traditional approaches showed considerably lower performance, with BioBERT models achieving F1-scores ranging from 0.536 to 0.802, substantially below the LLM performance range of 0.582 to 0.948. This comprehensive comparison confirms the consistent superiority of LLMs across different architectures, datasets, and sentiment categories, validating the effectiveness of expert knowledge integration approaches for health-related SA.

\begin{table*}[!t]
\caption{Performance of models, measured by Precision, Recall and F1-score. The highest value in each column on each data set is in bold.}
\label{tab:performance}
\centering
\scriptsize
\begin{tabular}{|l|ccc|ccc|ccc|}
\hline
\multirow{2}{*}{Model} & \multicolumn{3}{c|}{Positive} & \multicolumn{3}{c|}{Negative} & \multicolumn{3}{c|}{Neutral} \\
\cline{2-10}
 & Precision & Recall & F1-Score & Precision & Recall & F1-Score & Precision & Recall & F1-Score \\
\hline
\multicolumn{10}{|c|}{\textit{AUK Data Set}} \\
\hline
GPT-4.1 & 0.939 & 0.886 & 0.912 & 0.719 & 0.852 & 0.780 & 0.757 & 0.683 & 0.718 \\
GPT-4.1-fs & \textbf{0.950} & 0.914 & 0.932 & 0.796 & 0.796 & 0.796 & 0.733 & \textbf{0.805} & \textbf{0.767} \\
GPT-4.1-mini & 0.947 & 0.857 & 0.900 & 0.719 & 0.852 & 0.780 & 0.707 & 0.707 & 0.707 \\
GPT-4.1-mini-fs & 0.934 & 0.943 & 0.938 & \textbf{0.837} & 0.759 & 0.796 & 0.711 & 0.780 & 0.744 \\
GPT-o3 & 0.917 & 0.952 & 0.935 & 0.706 & \textbf{0.889} & 0.787 & \textbf{0.913} & 0.512 & 0.656 \\
GPT-o3-fs & 0.927 & \textbf{0.971} & \textbf{0.949} & 0.780 & 0.852 & \textbf{0.814} & 0.839 & 0.634 & 0.722 \\
GPT-o3-mini & 0.948 & 0.876 & 0.911 & 0.734 & 0.870 & 0.797 & 0.718 & 0.683 & 0.700 \\
GPT-o3-mini-fs & 0.939 & 0.876 & 0.906 & 0.719 & 0.852 & 0.780 & 0.789 & 0.732 & 0.759 \\
BioBERT-Balance & 0.914 & 0.505 & 0.650 & 0.489 & 0.852 & 0.622 & 0.458 & 0.537 & 0.494 \\
BioBERT-Combine & 0.894 & 0.562 & 0.690 & 0.674 & 0.537 & 0.598 & 0.341 & 0.756 & 0.470 \\
BioBERT-Covid & 0.903 & 0.533 & 0.671 & 0.500 & 0.759 & 0.603 & 0.464 & 0.634 & 0.536 \\
BioBERT-SemEval & 0.903 & 0.533 & 0.671 & 0.500 & 0.759 & 0.603 & 0.464 & 0.634 & 0.536 \\
SentiWordNet & 0.569 & 0.667 & 0.614 & 0.356 & 0.389 & 0.372 & 0.278 & 0.122 & 0.169 \\
TextBlob & 0.630 & 0.714 & 0.670 & 0.400 & 0.333 & 0.364 & 0.306 & 0.268 & 0.286 \\
VADER & 0.681 & 0.752 & 0.715 & 0.532 & 0.611 & 0.569 & 0.364 & 0.195 & 0.254 \\
\hline
\multicolumn{10}{|c|}{\textit{BLF Data Set}} \\
\hline
GPT-4.1 & 0.934 & 0.832 & 0.880 & \textbf{0.889} & \textbf{0.970} & \textbf{0.928} & 0.476 & 0.667 & 0.556 \\
GPT-4.1-fs & 0.925 & 0.905 & 0.915 & 0.882 & 0.909 & 0.896 & 0.562 & 0.600 & 0.581 \\
GPT-4.1-mini & 0.951 & 0.847 & 0.896 & 0.780 & \textbf{0.970} & 0.865 & 0.514 & 0.633 & 0.567 \\
GPT-4.1-mini-fs & \textbf{0.958} & 0.839 & 0.895 & 0.833 & 0.909 & 0.870 & 0.477 & \textbf{0.700} & 0.568 \\
GPT-o3 & 0.886 & \textbf{0.964} & \textbf{0.923} & 0.818 & 0.818 & 0.818 & \textbf{0.778} & 0.467 & \textbf{0.583} \\
GPT-o3-fs & 0.900 & 0.920 & 0.910 & 0.857 & 0.909 & 0.882 & 0.640 & 0.533 & 0.582 \\
GPT-o3-mini & 0.951 & 0.854 & 0.900 & 0.738 & 0.939 & 0.827 & 0.514 & 0.600 & 0.554 \\
GPT-o3-mini-fs & 0.938 & 0.876 & 0.906 & 0.806 & 0.879 & 0.841 & 0.500 & 0.600 & 0.545 \\
BioBERT-Balance & 0.924 & 0.708 & 0.802 & 0.519 & 0.848 & 0.644 & 0.488 & 0.667 & 0.563 \\
BioBERT-Combine & 0.894 & 0.737 & 0.808 & 0.692 & 0.545 & 0.610 & 0.295 & 0.600 & 0.396 \\
BioBERT-Covid & 0.949 & 0.679 & 0.791 & 0.500 & 0.818 & 0.621 & 0.417 & 0.667 & 0.513 \\
BioBERT-SemEval & 0.949 & 0.679 & 0.791 & 0.500 & 0.818 & 0.621 & 0.417 & 0.667 & 0.513 \\
SentiWordNet & 0.754 & 0.737 & 0.745 & 0.289 & 0.394 & 0.333 & 0.048 & 0.033 & 0.039 \\
TextBlob & 0.630 & 0.714 & 0.670 & 0.400 & 0.333 & 0.364 & 0.306 & 0.268 & 0.286 \\
VADER & 0.681 & 0.752 & 0.715 & 0.532 & 0.611 & 0.569 & 0.364 & 0.195 & 0.254 \\
\hline
\end{tabular}
\end{table*}

\begin{table*}[!t]
\caption{Performance comparison of best models in each category, measured by Precision, Recall and F1-score. The highest value in each column on each dataset is in bold.}
\label{tab:best_performance}
\centering
\scriptsize
\begin{tabular}{|l|ccc|ccc|ccc|}
\hline
\multirow{2}{*}{Model} & \multicolumn{3}{c|}{Positive} & \multicolumn{3}{c|}{Negative} & \multicolumn{3}{c|}{Neutral} \\
\cline{2-10}
 & Precision & Recall & F1-Score & Precision & Recall & F1-Score & Precision & Recall & F1-Score \\
\hline
\multicolumn{10}{|c|}{\textit{AUK Dataset}} \\
\hline
GPT-o3-fs & 0.927 & \textbf{0.971} & \textbf{0.949} & 0.780 & 0.852 & 0.814 & 0.839 & 0.634 & 0.722 \\
BioBERT-Covid & 0.903 & 0.533 & 0.671 & 0.500 & 0.759 & 0.603 & 0.464 & 0.634 & 0.536 \\
LLaMA 3.1-405B-fs & \textbf{0.935} & 0.962 & 0.948 & \textbf{0.880} & 0.815 & \textbf{0.846} & 0.738 & \textbf{0.756} & \textbf{0.747} \\
DeepSeek V3-fs & 0.933 & 0.933 & 0.933 & 0.746 & \textbf{0.926} & 0.826 & \textbf{0.857} & 0.585 & 0.696 \\
\hline
\multicolumn{10}{|c|}{\textit{BLF Dataset}} \\
\hline
GPT-o3-fs & 0.900 & 0.920 & 0.910 & 0.857 & 0.909 & 0.882 & 0.640 & 0.533 & 0.582 \\
BioBERT-Balance & 0.924 & 0.708 & 0.802 & 0.519 & 0.848 & 0.644 & 0.488 & 0.667 & 0.563 \\
LLaMA 3.1-405B-fs & \textbf{0.935} & \textbf{0.962} & \textbf{0.948} & \textbf{0.880} & 0.815 & \textbf{0.846} & \textbf{0.738} & \textbf{0.756} & \textbf{0.747} \\
DeepSeek V3-fs & 0.933 & 0.933 & 0.933 & 0.746 & \textbf{0.926} & 0.826 & 0.857 & 0.585 & 0.696 \\
\hline
\end{tabular}
\end{table*}

\begin{figure}[H]
\centerline{\includegraphics[width=0.85\columnwidth]{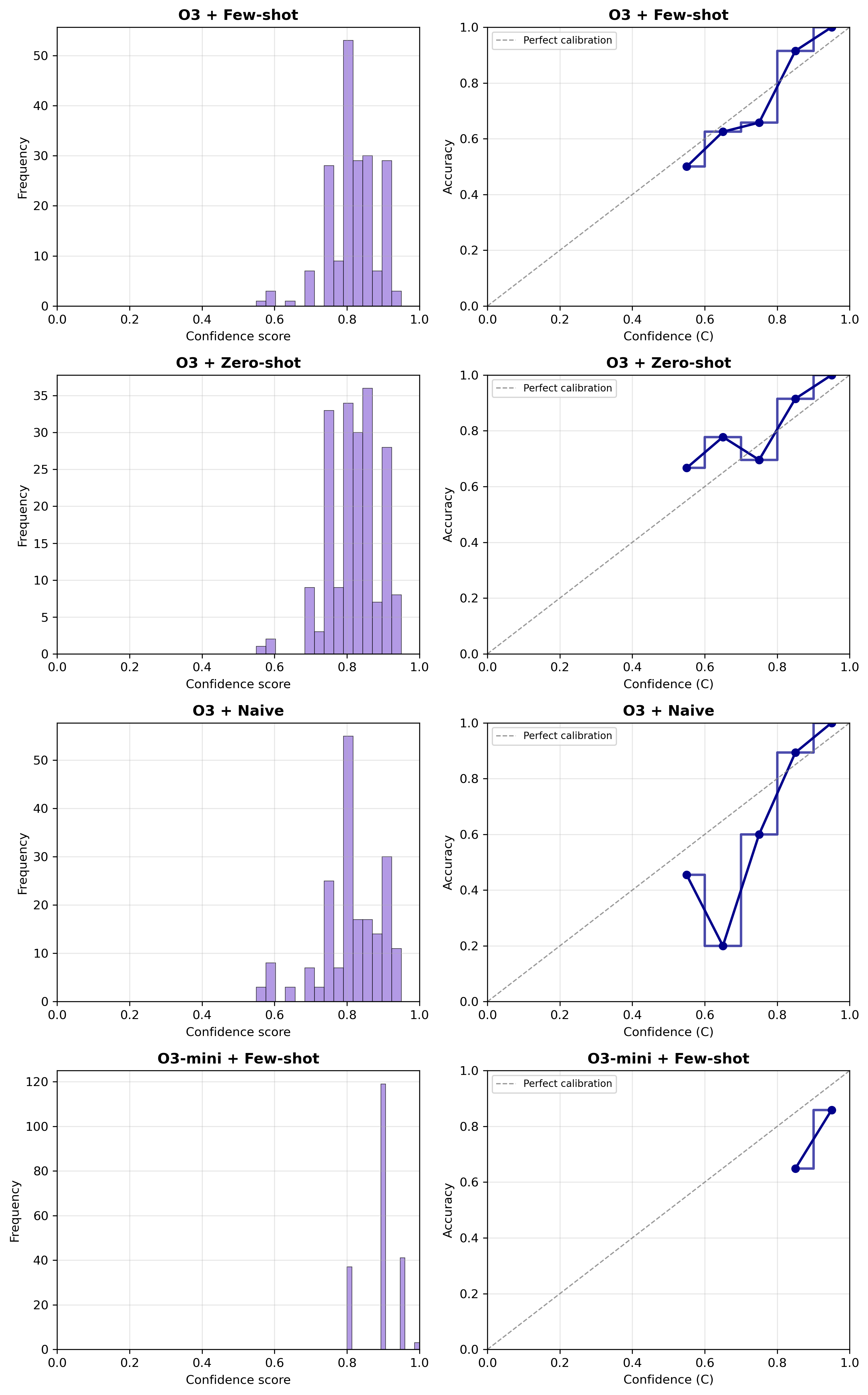}}
\caption{Confidence calibration analysis across different model-prompt combinations. The left column displays confidence score distributions, while the right column shows calibration curves (predicted confidence vs. actual accuracy). Diagonal lines represent perfect calibration.}
\label{fig:confidence}
\end{figure}

Concerning the last key point, Figure~\ref{fig:confidence} presents confidence calibration results for o3 model variants across different prompting strategies and o3-mini with few-shot prompting as a representative comparison. Due to space constraints, GPT-4.1 results are summarised textually below. The o3 model demonstrates superior confidence estimation with well-distributed scores between 0.5 and 1, enabling effective prediction quality differentiation.
Few-shot o3 model emerges as the optimal strategy, showing a strong linear relationship between confidence and accuracy, with the calibration curve closely tracking the perfect calibration line. This linear progression indicates that higher confidence scores reliably correspond to higher accuracy, making confidence a valuable indicator for automated label quality assessment. In contrast, zero-shot prompting shows more variable calibration patterns, while naive prompting (i.e., no instruction from the codebook) exhibits notable deviations from perfect calibration, particularly in mid-confidence ranges.
The o3-mini few-shot results shown are representative of all o3-mini configurations due to consistent patterns across prompting strategies. o3-mini models show severely concentrated confidence distributions above 0.8 across all prompting strategies, rendering confidence scores uninformative for quality assessment. Prompting strategy has minimal impact on o3-mini due to fundamental limitations in confidence estimation. Similarly, all GPT-4.1 variants exhibited limitations similar to o3-mini. Unlike o3 models, GPT-4.1 variants are not specifically designed for reasoning tasks, which may explain their inability to produce meaningful confidence estimates.
The superior performance of o3 models with few-shot prompting suggests that reasoning capabilities, combined with contextual examples, enable better understanding of confidence semantics, translating into meaningful uncertainty estimates. These results indicate that only o3 models with few-shot prompting provide confidence scores suitable for label quality evaluation and selective prediction strategies.

\section{Discussion}
\label{sec:discussion}

\subsection{Principal Results}

This study demonstrates that LLMs can successfully integrate expert knowledge to achieve sophisticated SA in digital health contexts without requiring extensive labelled training data. Our systematic evaluation reveals that LLMs consistently outperformed traditional approaches across multiple performance metrics, with the most significant finding being their ability to achieve expert-level agreement ($\kappa$: 0.42-0.75) through structured knowledge integration rather than data-intensive training.

The core methodological innovation lies in our codebook-guided approach, which systematically transfers expert domain knowledge through structured prompting. Unlike retrieval-augmented generation (RAG) methods that dynamically retrieve information from external knowledge bases, our approach embeds expert-derived interpretation guidelines directly into prompts, enabling consistent application of domain-specific rules. 

This approach is particularly effective for SA where classification rules can be systematically defined and consistently applied across datasets, representing a fundamental shift from traditional data-intensive methods to knowledge-guided approaches in healthcare analytics. This paradigm shift directly addresses fundamental limitations of existing methods in healthcare contexts. While lexicon-based models lack the contextual understanding required for health-related content, and traditional pre-trained models like BioBERT require substantial fine-tuning with labelled datasets that are scarce in healthcare due to privacy constraints, LLMs demonstrated the ability to apply expert-derived guidelines directly without domain-specific training.

The practical implications extend beyond performance improvements to address critical barriers in healthcare analytics: the high cost of expert annotation and the technical expertise traditionally required for advanced text analysis. By enabling sophisticated analysis through structured prompting rather than complex model development, this approach democratizes access to expert-quality text interpretation, making advanced analytics feasible for healthcare research teams regardless of their technical infrastructure or ML expertise. This accessibility is particularly valuable for smaller healthcare organisations and research teams with limited resources, where the ability to achieve expert-quality results through knowledge integration rather than extensive data collection reduces both financial burdens and technical barriers. The structured codebook methodology offers distinct advantages for healthcare applications where expert interpretation rules can be systematically defined and consistently applied across large datasets.

Our confidence calibration analysis reveals important distinctions in uncertainty estimation capabilities across LLM architectures. The GPT-o3 model with few-shot prompting demonstrated superior confidence calibration, showing a strong linear relationship between predicted confidence and actual accuracy. This calibration quality enables practical deployment strategies where confidence scores can reliably indicate prediction quality, supporting automated processing of high-confidence cases while flagging uncertain predictions for expert review.

In contrast, other model variants, including GPT-4.1 and GPT-o3-mini, showed concentrated confidence distributions that limit their utility for quality assessment. These findings suggest that confidence estimation capabilities vary significantly across model architectures and training approaches, with reasoning models like GPT-o3 better equipped to provide meaningful uncertainty estimates. This has important implications for healthcare deployment, where reliable confidence estimation is crucial for maintaining quality assurance while maximising automation efficiency.

The consistency of results across different LLM architectures—including GPT models, LLaMA, and DeepSeek—validates the robustness of this knowledge integration approach and provides healthcare organisations with crucial implementation flexibility. This cross-architectural validation is particularly important for healthcare applications, where open-source models like DeepSeek and LLaMA offer significant advantages for local deployment, allowing organisations to select models based on their specific security, privacy, and resource requirements.

Our findings have significant implications for digital health applications beyond SA. The ability to encode expert knowledge into structured guidelines that LLMs can reliably apply offers a scalable solution for various healthcare text analysis tasks, from clinical documentation interpretation to patient experience monitoring. The demonstrated effectiveness in OHCs, with their complex emotional expressions and medical terminology, suggests strong potential for application in other challenging healthcare text analysis contexts. The structured knowledge integration framework extends beyond SA to other expert annotation tasks in healthcare, providing a generalisable approach for scaling domain expertise across diverse text analysis applications. This knowledge integration approach offers a practical pathway for implementing advanced analytics in resource-constrained healthcare environments whilst maintaining expert-quality interpretive standards.

\subsection{Strengths and Limitations}

A key strength of this study lies in its comprehensive evaluation design, systematically comparing LLMs against both traditional ML and lexicon-based approaches using expert-annotated healthcare data. The rigorous annotation process involving five experts with PhD qualifications, including a clinician, ensured high-quality gold standard labels for meaningful performance assessment. The inclusion of both zero-shot and few-shot learning evaluation provides practical insights into LLM deployment scenarios where training data availability varies.

The focus on OHCs as a challenging test case represents another strength, as OHC content exhibits complex sentiment patterns, medical terminology, and emotional nuance that extend beyond typical social media analysis. This context provides a robust evaluation environment for assessing automated approaches in healthcare-specific settings where traditional methods often struggle.

From a methodological perspective, the structured codebook approach offers a replicable framework for knowledge transfer across different healthcare domains and research contexts. This systematic method for expert knowledge encoding provides a practical roadmap for researchers seeking to implement similar analysis capabilities without requiring extensive ML expertise. The evaluation of multiple LLM architectures, including open-source models, provides implementation flexibility for organisations with different security and privacy requirements, particularly important when handling confidential medical data. 

Automated SA through expert knowledge integration enables large-scale monitoring of patient discussions, offering insights into treatment responses and peer support effectiveness that would otherwise require extensive expert review. This approach enables access to sophisticated analytical capabilities, enabling healthcare research teams to conduct advanced text analysis without requiring extensive technical infrastructure. Our open-source implementation further enhances accessibility by providing ready-to-use tools that enable healthcare researchers to apply these advanced analytical capabilities immediately, without the complexity of model training, fine-tuning, or extensive technical infrastructure typically required for traditional ML approaches.

The confidence calibration analysis provides additional practical value by identifying models suitable for quality-controlled deployment scenarios. The demonstrated linear relationship between confidence and accuracy in GPT-o3 few-shot models enables the implementation of selective prediction strategies, where high-confidence predictions can be processed automatically while uncertain cases receive human oversight. This capability addresses a critical need in healthcare applications where balancing automation efficiency with quality assurance is paramount.

This study also has several limitations that warrant consideration. Our evaluation focused primarily on SA within OHCs, and the broader applicability of the structured prompting approach across different digital health text analysis contexts remains to be established. The knowledge integration methodology, whilst demonstrating strong performance, depends heavily on the quality and comprehensiveness of expert-derived guidelines. Our codebook development involved extensive consensus-building among five expert annotators, but this process may be challenging to replicate across different healthcare domains or cultural contexts where expert perspectives might vary significantly. Standardisation approaches for codebook development across different types of medical expertise require further investigation.

Additionally, while we evaluated multiple LLM architectures to demonstrate robustness, our assessment was limited to zero-shot and few-shot learning scenarios. The potential benefits of fine-tuning LLMs specifically for healthcare applications, particularly with privacy-preserving techniques, represent an important area for future investigation. The trade-offs between knowledge-guided prompting and domain-specific fine-tuning require further exploration.

Furthermore, although our statistical validation confirmed expert-level agreement, the deployment of LLMs in clinical or high-stakes healthcare contexts requires careful consideration of interpretability, accountability, and ethical implications. SA in healthcare often involves subjective judgement and contextual nuance that may require ongoing human oversight, particularly in settings where patient safety or clinical decision-making are directly involved.

While our structured prompting approach proved effective for SA, future research could investigate how hybrid methodologies combining our codebook-guided approach with RAG mechanisms might enhance performance for more complex analytical tasks requiring dynamic access to evolving medical knowledge or real-time information integration. Additionally, exploring the application of this knowledge integration framework across different digital health text analysis contexts would further validate the generalisability of the approach.

\section{Conclusions}
\label{sec:conclusions}

This study demonstrates that LLMs can effectively integrate expert knowledge to achieve sophisticated SA in digital health contexts through structured prompting approaches. Our systematic evaluation establishes the feasibility of knowledge-guided analysis that achieves expert-quality results without requiring extensive training data. Reasoning models also demonstrated reliable uncertainty estimation capabilities, supporting deployment strategies suitable for healthcare applications.

The core contribution lies in demonstrating how expert domain knowledge can be systematically transferred to automated systems through structured knowledge integration, enabling sophisticated healthcare text analysis without traditional data requirements. This approach offers implementation flexibility across different technological contexts while maintaining analytical quality standards.

As healthcare increasingly relies on data-driven insights, LLMs offer a transformative pathway for scaling expert knowledge across diverse digital health applications. Our open-source implementation facilitates immediate adoption across healthcare research contexts, with demonstrated uncertainty estimation capabilities that further support practical deployment. The framework enables researchers to readily test its applicability to other digital health text annotation tasks, facilitating rapid validation and subsequent deployment. 

Future research should explore the application of knowledge integration frameworks across diverse digital health contexts to further validate the generalisability of this approach and support the development of more accessible, expert-quality analytical tools for healthcare research and practice.

\section*{Ethical Considerations}
\label{sec:ethics}

The study was approved by Queen Mary University of London's Ethics of Research Committee (QMERC22·279). In addition, the research protocol was examined and permission to undertake the research was obtained from AUK, BLF charities, as well as HealthUnlocked.

\section*{Acknowledgment}

The authors would like to thank Asthma UK, British Lung Foundation and HealthUnlocked for granting the permission to conduct the study.

\bibliographystyle{IEEEtran}
\bibliography{references}

\end{document}